\documentclass{article}

\usepackage{microtype}
\usepackage{graphicx}
\usepackage{subfigure}
\usepackage{booktabs} 

\usepackage{hyperref}

\usepackage[accepted]{icml2025}

\usepackage{amsmath}
\usepackage{amssymb}
\usepackage{mathtools}
\usepackage{amsthm}
\usepackage{caption}
\usepackage{tabularx}
\usepackage[capitalize,noabbrev]{cleveref}
\usepackage{wrapfig}
\usepackage{ragged2e}
\usepackage{enumitem}
\setlist[enumerate]{itemsep=0pt, topsep=0pt, leftmargin=*}
\theoremstyle{plain}

\theoremstyle{definition}

\theoremstyle{remark}

\usepackage[textsize=tiny]{todonotes}

\icmltitlerunning{A note on data-efficient finetuning of vision foundation models}

\begin{document}

\twocolumn[
\icmltitle{A note on data-efficient finetuning of vision foundation models}



\icmlsetsymbol{equal}{*}

\begin{icmlauthorlist}
\icmlauthor{Niclas Hergenröther}{ellis}
\icmlauthor{Antonio Orvieto}{ellis}
\end{icmlauthorlist}
\icmlcorrespondingauthor{Niclas Hergenröther}{niclas.hergenroether@student.uni-tuebingen.de}
\icmlaffiliation{ellis}{ELLIS Institute Tübingen, Max Planck Institute for Intelligent Systems, Tübingen AI
Center}

\icmlkeywords{Machine Learning, Visual Prompting, VP, Transfer Learning, Adversarial Reprogramming}

\vskip 0.3in
]



\printAffiliationsAndNotice{} 

\begin{abstract}
Adapting foundation models to a new task within or beyond the training domain is the de facto standard for building specialized machine learning models. Recently, Visual Prompting (VP) has emerged as a promising technique for efficient knowledge transfer. As new foundation model families~(like Mamba) get introduced and VP pipelines such as AutoVP reach greater maturity, we find a growing need for a systematic evaluation of current approaches. In this work, we assess the performance of the latest models, comparing them to earlier architectures and alternative fine-tuning methods, to better understand the progress, challenges and opportunities in the field of efficient fine-tuning under resource limitations. Towards this goal, this paper provides a concise empirical overview of the interactions among foundation model families (Attention-, Convolution-, and Mamba-based) and transfer paradigms: VP, Linear Probing (LP), Full Fine-tuning (FFT) and Low Rank Adaptation (LoRA). Our work builds up on previous findings by broadening the selection of evaluated models and techniques while tuning hyperparameters. In the interest of delivering practical guidelines for the user, we also explore application of prevalent regularization techniques to boost performance in the context of VP.
\end{abstract}
\vspace{-9mm}
\section{Introduction}
\vspace{-2mm}
Efficient transfer of pretraining knowledge to new tasks and domains is an essential requirement for successful downstream applicability of foundation models. As the AI landscape continues to grow -- spanning diverse architectures and design choices -- so do options for fine-tuning, including new pipelines for Visual Prompting (VP), Linear Probing (LP), Full Fine-tuning (FFT) and Low Rank Adaptation (LoRA).\\
For a review of these methods, please refer to Section~\ref{sec:background}. Given this rapid expansion, it is critical to systematically and regularly assess how different architectural choices interact with various transfer learning methods. In this work, we present an empirical evaluation of VP across several prominent model architectures, analyze how performance correlates with pretraining accuracy, and discuss potential performance and compute tradeoffs when comparing it to other fine-tuning methods on challenging tasks.

\begin{figure}
    \centering
    \includegraphics[width=\linewidth]{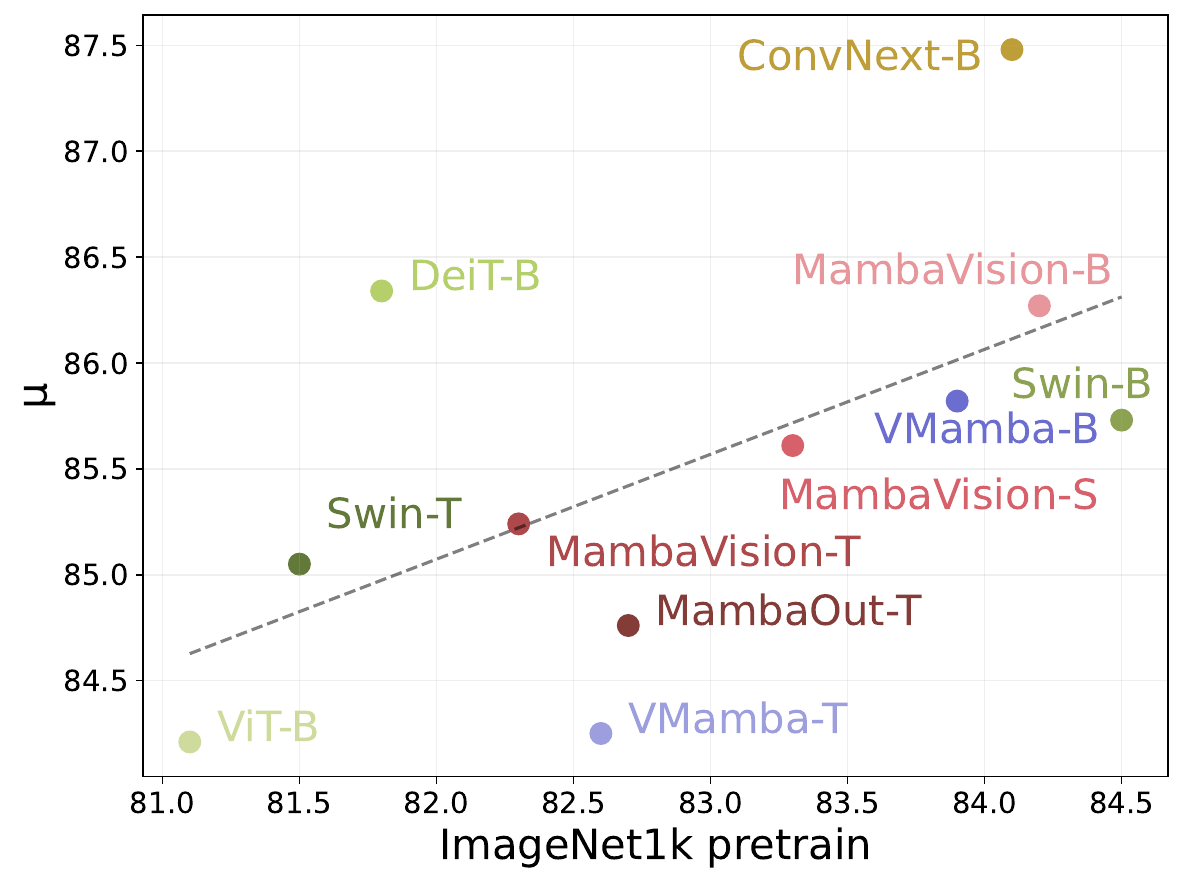}
    \caption{\textit{ImageNet1k-pretrain and average VP top-1 accuracy over the tasks described in Section~\ref{sec:setup}. We follow the recipe of~\citet{tsao2023autovp}. All models are tuned to best. Dotted line is linear regression, full results in Table~\ref{tab:VP Results}. }}
    \label{fig:ImageNet mu corr}
    \vspace{-6mm}
\end{figure}

\begin{table*}[h]
    \centering
    \begin{sc}
    \begin{tabularx}{\linewidth}{|X|c c c c c c c c c | c|c|}
    \hline
    \centering{} & \rotatebox{90}{CIFAR10} & \rotatebox{90}{CIFAR100} & \rotatebox{90}{Flowers102} & \rotatebox{90}{GTSRB} & \rotatebox{90}{DTD} & \rotatebox{90}{EuroSAT} & \rotatebox{90}{Food101} & \rotatebox{90}{OxfordPets} & \rotatebox{90}{$\mu$} & \rotatebox{90}{pretrain acc.} \\
    \hline
    VMamba-T                & 94.89  & 80.53  & 91.67  & 86.75  & 67.17  & 95.72  & 69.25  & 88.07  & 84.25  & 82.60  \\ 
    MambaOut-T              & 95.07  & 81.36  & 89.58  & 88.27  & 67.73  & 95.70  & 71.26  & 89.14  & 84.76  & 82.70  \\ 
    MambaVision-T           & 95.44  & 80.89  & 92.55  & 87.74  & 66.26  & 96.56  & 71.56  & 90.91  & 85.24  & 82.30  \\ 
    Swin-T                  & 94.28  & 80.32  & 92.34  & 88.95  & 66.37  & 96.31  & 72.99  & 88.83  & 85.05  & 81.50  \\
    MambaVision-S           & 95.97  & 82.37  & 90.71  & 86.93  & 66.58  & 96.62  & 74.18  & 91.57  & 85.61  & 83.30  \\ 
    \hline
    MambaVision-B           & 96.54  & 83.88  & 91.79  & 88.79  & 65.64  & 96.63  & \textbf{75.65}  & 91.22  & 86.27  & 84.20  \\ 
    ConvNext-B              & 96.42  & \textbf{84.44}  & \textbf{94.15}  & 90.28  & \textbf{70.77}  & 96.61  & 75.10  & \textbf{92.04}  & \textbf{87.48}  & 84.10  \\ 
    VMamba-B                & \textbf{96.58}  & 76.90  & 93.67  & 90.56  & 67.87  & 94.14  & 75.52  & 91.35  & 85.82  & 83.90  \\
    DeiT-B	                & 96.52	 & 83.59  &	93.51  & \textbf{91.57}	& 66.54	 & \textbf{96.85}  &	72.13  & 90.01  & 86.34	 & 81.80  \\
    ViT-B                   & 95.52  & 78.18  & 90.74  & 89.54  & 62.13  & 96.80  & 70.84  & 89.97  & 84.21  & 81.10  \\
    Swin-B                  & 95.45  & 81.55  & 93.35  & 86.19  & 68.47  & 96.59  & 74.26  & 89.95  & 85.73  & 84.50  \\\hline
    Corr. ImageNet          & 0.49	 & 0.31	  & 0.35   & -0.23	& 0.66	 & -0.23  & 0.73   & 0.53	& 0.60   &        \\ \hline
    \end{tabularx}
    \end{sc}
     \caption{\textit{Top-1 accuracy after training with AutoVP using fully connected label mapping. $\mu$ denotes the mean fine-tuning performance. The last column represents the ImageNet pretraining accuracy. The bottom row reports the correlation coefficient between the corresponding column and the ImageNet pretraining accuracy.}}
    \label{tab:VP Results}
    \vspace{-6mm}
\end{table*}

\begin{enumerate}[label=\arabic*., itemsep=0pt, topsep=0pt, leftmargin=*]
    \item Compared to earlier results~\citep{misra2024lowshottransferabilityvmamba} for ILM-VP~\cite{chen2023understanding}, we found that vision transformers and Mamba-based \cite{gu2023mamba} models perform similarly -- with only a slight edge for the former -- under latest VP practice~\citep{tsao2023autovp}. However, when selecting a model for VP, \textbf{ConvNext appears to be the best choice} from our selection~(Figure~\ref{fig:ImageNet mu corr}). References and a short description for these models are presented in Section~\ref{sec:setup}.
    \vspace{-1mm}
    \item \textbf{On DTD, LoRA achieves higher reported accuracy than FFT for five of the six evaluated
    models while training fewer parameters.}. State-of-the-art VP practice can underperform~(overfit) on data scarce tasks compared to LP, FFT and LoRA -- see Table~\ref{tab:VP_LP_FFT}.  In settings with restricted computational resources, LP seems to be dominant w.r.t. VP.
    \vspace{-1mm}
    \item It is natural to ask at this point if VP performance can be improved via regularization. We found that prevalent \textbf{regularization techniques can boost generalization}, whilst still underperforming LoRA. We noticed that most gains can be observed when applying regularization to the label mapping weights, as opposed to the prompt.
\end{enumerate}

\section{Background}
\label{sec:background}
\vspace{-2mm}
\paragraph{Benchmarking Visual Prompting.} VP, first introduced as Adversarial Reprogramming \cite{elsayed2018adversarialreprogrammingneuralnetworks}, leverages the idea of utilizing a \textbf{universal input perturbation} to transfer a pretrained model to a new domain or task. The term Prompting originates from language modeling \cite{lester2021power} and was later adapted to the vision domain \cite{bahng2022exploring}. VP keeps the pretrained source model entirely frozen and therefore differs from other techniques~(see next paragraph) that may require adapting source model weights. Black-box variants of reprogramming can also work with API-only access \cite{tsai2020transferlearningknowingreprogramming}.

With AutoVP, \citet{tsao2023autovp} introduced improvements of the VP methodology as well as a unified benchmark for VP. The introduced improvements include the possibility of resizing the prompt as well as a new Label Mapping (LM) technique: Fully Connected Label Mapping (FCLM). FCLM is realized by a fully connected layer of dimensions $C_S \times C_T$ (\# of source classes $C_S$;\# of target classes $C_T$).

While there exists research on the combination of new foundation models and VP \cite{chen2023understanding, tsao2023autovp, misra2024lowshottransferabilityvmamba}, these works are either limited in model selection or didn't tune hyperparameters, making the results ambiguous to interpret.
In this work, we evaluate a selection of ImageNet1k pretrained foundation models (Sec. \ref{sec:setup}) using AutoVP \cite{tsao2023autovp} on a common set of downstream image-classification tasks and in addition to performance (Tab. \ref{tab:VP Results} and Fig. \ref{fig:ImageNet mu corr}) also investigate on the behavior of models w.r.t. the resizing of the prompt (Tab. \ref{tab:VP Prompt size} and Fig. \ref{fig:Resize}).
\vspace{-2mm}
\begin{table}[h]
    \centering
    \begin{sc}
    \begin{tabular}{|l|cccc|}
        \hline
                        &  VP   & LP    & FFT               & LoRA  \\
        \hline
        MambaVision     & 65.64 & 70.21 & \textbf{75.16 }   & 73.88\\
        ConvNext        & 70.77 & 70.90 & 73.18             & \textbf{76.60}\\
        VMamba          & 67.87 & 67.45 & 73.30             & \textbf{75.00}\\
        DeiT            & 66.54 & 68.56 & 72.98             & \textbf{74.68}\\       
        ViT             & 62.13 & 71.38 & 75.32             & \textbf{77.02}\\
        Swin            & 68.47 & 77.07 & 79.63             & \textbf{81.33}\\
        \hline
    \end{tabular}
    \end{sc}
    \caption{\textit{Top 1 test-performance in \% of various base-models on DTD using different Transfer Learning paradigms. Learning rate, rank (LoRA) and weight decay tuned. Base-size used for all models.}}
    \label{tab:VP_LP_FFT}
    \vspace{-4mm}
\end{table}
\paragraph{Full fine-tuning and linear probing.} Based on the improvements of AutoVP, we conduct a comparison of FFT, LP, LoRA ~(described next) and VP on the data scarce DTD dataset~\cite{6909856}, which showed to be the most challenging task in the benchmark. We investigate a subset of the models evaluated in Tab. \ref{tab:VP Results}, while tuning learning rate and weight decay to ensure optimal performance for each method. LP freezes the pretrained model, whilst only replacing and training a new classifier with the idea of leveraging the existing feature representations at the penultimate layer of a model~\cite{alain2018understandingintermediatelayersusing}. In contrast to full fine-tuning, LP drastically reduces the amount of trainable parameters as well as hardware requirements. Last, in our implementation of FFT, we account for the new number of output classes by training a new classifier from scratch \cite{chen2023understanding}. In contrast to LP or VP, no model parameters are frozen. While this unlocks stronger results, it also requires larger compute and memory~(Tab.\ref{tab:param_comp}) compared to the other approaches.
\vspace{-4mm}
\paragraph{LoRA.} Low-Rank Adaptation (LoRA) \cite{hu2022lora} is one of the most prominent lines of PEFT research. Instead of updating the weight matrix $W \in \mathbb{R}^{d \times k}$ directly, LoRA parametrizes the update as a product $\Delta W = BA$, of low-rank $B \in \mathbb{R}^{d \times r}$, $A \in \mathbb{R}^{r \times k}$ and hyperparameter $r \ll \min(d, k)$. Only $A$ and $B$ are trained, which drastically reduces the number of trainable parameters. Since the LoRA modules contain only these low-rank updates and not the base model's weights, they are lightweight and can be swapped depending on the task, while the base model remains frozen and unchanged.

\vspace{-4mm}
\paragraph{Regularizing Visual Prompting.} We noticed models showed signs of overfitting when being transferred to DTD using VP -- as testified by Table~\ref{tab:VP_LP_FFT}, where \textbf{all models reached $\mathbf{\geq 99\%}$ training accuracy despite suboptimal test performance compared to LP, FFT or LoRA}. This poses the question on how to effectively regularize VP. This task is particularly interesting, as common regularization methods are motivated by a different learning paradigm. We investigated a set of common regularization approaches while also ablating on targeting regularizing only to a subset of the optimized parameters in VP: either prompt or Label Mapping parameters. Specifically, we evaluated performance of Dropout \cite{hinton2012improving}, L1-/L2-regularization, Sharpness-aware minimization~(SAM) \cite{foret2020sharpness} and Autoaugment \cite{cubuk2019autoaugment}. 
\section{Experimental setup}\label{sec:setup}
\vspace{-2mm}
The tasks are selected in accordance with earlier evaluations on VP \cite{chen2023understanding, tsao2023autovp}, consisting of CIFAR10/100 \cite{Krizhevsky2009LearningML} Flowers102 \cite{Nilsback2008AutomatedFC},GTSRB\cite{stallkamp2011german}, DTD \cite{6909856}, EuroSAT \cite{helber2019eurosatnoveldatasetdeep}, Food101 \cite{bossard14}, OxfordPets \cite{Parkhi2012CatsAD}.\\
The models we use are VMamba\footnotemark[1] \cite{liu2024vmambavisualstatespace}, MambaOut\footnotemark[2] \cite{yu2024mambaoutreallyneedmamba}, MambaVision\footnotemark[3] \cite{hatamizadeh2024mambavision}, SwinT\footnotemark[4] \cite{liu2021swintransformerhierarchicalvision}, ConvNext\footnotemark[4] \cite{liu2022convnet2020s}, ViT\footnotemark[4] \cite{dosovitskiy2020image} and DeiT\footnotemark[5]\cite{touvron2021trainingdataefficientimagetransformers}.
In all cases, the ImageNet \cite{deng2009imagenet} pretrained weights are utilized.
\footnotetext[1]{VMamba repository (\href{https://github.com/MzeroMiko/VMamba}{github.com/MzeroMiko/VMamba)}}
\footnotetext[2]{MambaOut repository (\href{https://github.com/yuweihao/MambaOut}{github.com/yuweihao/MambaOut)}}
\footnotetext[3]{MambaVision repository (\href{https://github.com/NVlabs/MambaVision}{github.com/NVlabs/MambaVision)}}
\footnotetext[4]{Imported using torchvision \cite{torchvision2016}}
\footnotetext[5]{Imported using timm \cite{rw2019timm}}
\begin{itemize}
    \item Our initial set of experiments~(Table~\ref{tab:VP Results} and Section~\ref{sec:VP}) involves training the previously introduced models -- belonging to drastically different architecture families -- on the AutoVP Benchmark~\citep{tsao2023autovp}. For detailed information regarding hyperparameter tuning, please refer to Appendix \ref{App: AutoVP}.
    \vspace{-2mm}
    \item Our second series of experiments~(Section~\ref{sec:paradigms}) entails a comparative evaluation of LP, VP, FFT and LoRA on a subset of the introduced models. These fine-tuning paradigms are compared on the DTD dataset, as it appears to be the most challenging task with limited training data.
    \vspace{-2mm}
    \item Our third set of experiments~(Section~\ref{sec:regularization}) focuses on the regularization of VP. To assess regularization strategies on VP, we evaluate VMamba-B on the DTD Dataset (with Vanilla-VP performance obtainable from Table \ref{tab:VP Results}), utilizing a fixed learning rate of $0.001$. Dropout and L1/L2 regularization are implemented both across the entire model or specifically targeted to either the mapping or the prompt parameters.
\end{itemize}
\vspace{-2mm}
All runs from the first and second set of experiments are using a single Nvidia A100 GPU. The remaining experiments are utilizing either Nvidia Tesla V100 or Nvidia Quadro RTX 6000 GPUs.

The evaluation of different fine-tuning techniques must consider the distinct computational requirements for training models of significant size. The precise calculation of trainable model parameters for the different approaches is contingent on numerous factors, including the selected model, input dimensions, and the number of target classes. For this comparison we assume VMamba-B being applied to the DTD dataset with 47 target classes. For VP, inputs are resized using a learnable prompt size, while for LP, FFT and LoRA they are resized to the input size $224 \times 224$. Furthermore, while the trainable parameters of all approaches depend on the number of target classes, VP is also reliant on the input dimensions and the number of source classes in the pretrained model while LP also depends on the feature dimension of the penultimate layer. for LoRA, a rank of 2 was used to compute the reported number of parameters, as the different ranks all yielded similar results and therefore rank 2 showed to be the most lightweight variant reaching good performance. It is noteworthy that the implementation for LoRA in combination with VMamba was probably less optimized compared to the implementation for transformer based models. 
Consequently, the reported numbers need to be interpreted within the context of the specific choice of pretrained model and task.
\section{Results}
\subsection{VP levels model performances}\label{sec:VP}
Tab. \ref{tab:VP Results} presents the model performances alongside the correlation between model performance and pretraining ImageNet accuracy. In general, raw ImageNet pretraining accuracy presents itself to be a determining factor for model performance, as validated by the reported correlation coefficient. This indicates an external validity of ImageNet-performance \citep{torralba2011unbiased}. Nonetheless, certain tasks exhibit weak or even negative correlation. 
\begin{figure}[h]
\centering
\includegraphics[width=0.9\linewidth]{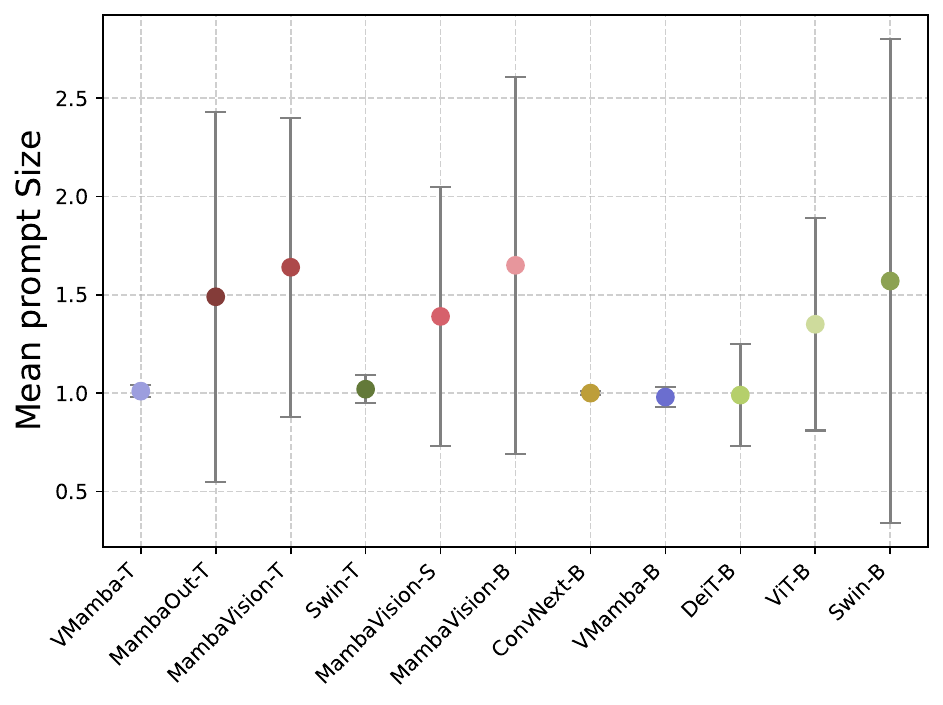}
\caption{\textit{Mean prompt size of the best performing run, with the initial prompt scaling being 1. Mean and standard deviation calculated over the tasks introduced in Sec. \ref{sec:setup}.}}
\label{fig:Resize}
\vspace{-3mm}
\end{figure}

ConvNext achieved the highest accuracy on four of the eight tasks. Yet, some tasks are more sensitive to model choice than others: while on CIFAR10 and EuroSAT performance differences are marginal, on DTD, the right selection of pretrained model is crucial. 
\begin{figure}[h]
\centering
\includegraphics[width=0.8\columnwidth]{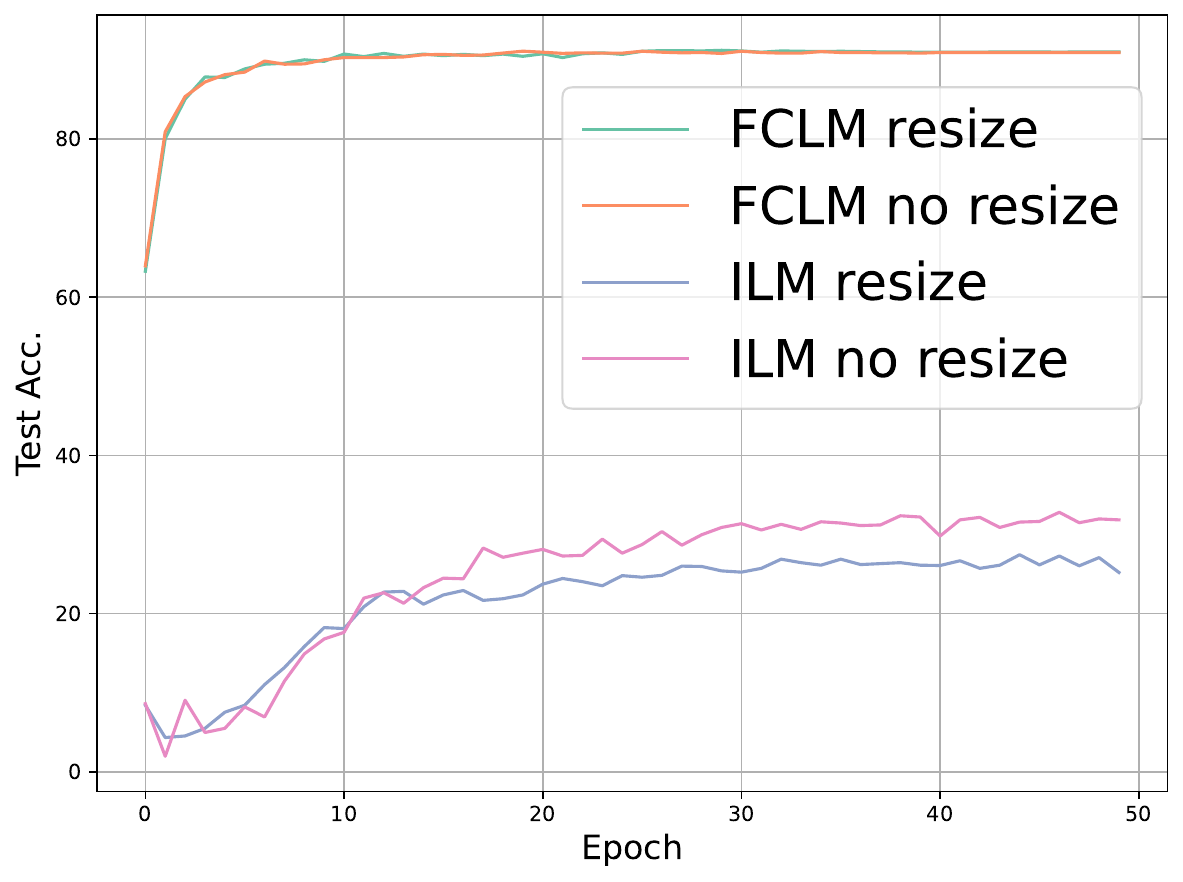}
\captionof{figure}{\textit{Flowers102 performance of VMamba-T for combinations of ILM and FCLM with and without enabled resizing.}}
\label{fig:App_resize}   
\vspace{-3mm}
\end{figure}
Our last investigation concerns the comparison with VP methods prior to~\citet{tsao2023autovp}, and in particular with ILM-VP~\citep{chen2023understanding}, tested also by~\citet{misra2024lowshottransferabilityvmamba}. In contrast to  ILM-VP, the variance in performance between different model architectures is significantly reduced. Using AutoVP on the Flowers102 dataset increases the performance of VMamba by 54.77 percentage points in comparison to ILM-VP, where it only reaches 38.9\%. In comparison, ViT-B already achieves 83.1\% under ILM-VP and is outperformed by VMamba under AutoVP.

For VMamba-T on Flowers102, our ablation separates the effects of label map-
ping and prompt resizing and identifies fully connected label mapping as the main source of the
improvement over ILM-VP (Fig. \ref{fig:App_resize}). As depicted in Fig. \ref{fig:Resize}, the capability to resize the prompt is not exploited uniformly across models.
ConvNext, which on average performs best, retains the initial configuration, while Swin-B exhibits considerable variation in prompt size. Models differing only by size (e.g., MambaVision, VMamba) tend to have similar behavior, while Swin Transformer behaves as an outlier with Swin-T maintaining the initial prompt size and Swin-B showing variation. Detailed results can be found in Appendix \ref{App: AutoVP}.
 \begin{table}[h]
 \begin{sc}
    \begin{tabularx}{\columnwidth}{|c|rXX|}
        \hline
                    & \# param.  & GPU-Mem (GB) & Epochtime (s)\\
        \hline
        VP          & 197,576    & 12.5         & 39.8 \\
        
        LP          & 48,175     & 11.2         & 36.1 \\
        
        FFT         & 87,601,967 & 72.3         & 49.2 \\
        
        LoRA        & 740,911    & 67.5         & 45.0 \\
        \hline
    \end{tabularx}
    \end{sc}
    \caption{\textit{VP refers to AutoVP with fully connected label mapping and adjustable prompt size. GPU-Mem describes the peak allocated GPU Memory. Reported Runtime is the median over all runs. All reported numbers refer to the use of VMamba-B on the DTD dataset.}}
    \label{tab:param_comp}
    \vspace{-2mm}
\end{table}
\vspace{-6mm}
\subsection{Comparison training paradigms}\label{sec:paradigms}
As showcased in Tab. \ref{tab:param_comp}, for the specific setting described above, LP proves to be the most lightweight fine-tuning method, with VP being only slightly more hardware demanding. By requiring only about 15\% of the GPU-memory required for FFT, the field of application is drastically broadened by, in contrast to FFT, being able to e.g. run on a Nvidia Tesla V100 GPU. \textbf{LoRA manages to outperform FFT on 5 of the 6 evaluated models, surprisingly not showing substantial differences between different rank configurations} (See tab. \ref{tab:LoRA r}). 
\vspace{-3mm}
\begin{table}[h]
    \centering
    \begin{sc}
    \begin{tabularx}{\columnwidth}{|l|XXXXX|}
        \hline
            \multicolumn{1}{|r}{rank $r$}  & \multicolumn{1}{c}{1}     &  \multicolumn{1}{c}{2}    & \multicolumn{1}{c}{4}     & \multicolumn{1}{c}{8}     & \multicolumn{1}{c|}{16}  \\
        \hline
        M-Vision        & 73.09 & 73.78 & 72.77 & 73.88 & \textbf{73.94} \\
        ConvNext        & 76.22 & 76.39 & 76.49 & \textbf{76.54} & 76.12\\
        VMamba          & 73.78 & \textbf{75.00} & 74.41 & 74.57 & 74.15\\
        DeiT            & 73.09 & 73.94 & \textbf{74.57} & 73.46 & 73.67\\       
        ViT             & 76.70 & 76.44 & 76.60 & 75.85 & \textbf{77.02}\\
        Swin            & 80.00 & \textbf{81.33} & 81.17 & 80.90 & 80.00\\
        \hline
    \end{tabularx}
    \end{sc}
    \caption{\textit{Top 1 test-performance in \% for different rank configurations.}}
    \label{tab:LoRA r}
    \vspace{-4mm}
\end{table}

It is also noteworthy, that whilst the majority of research conducted on LoRA is in combination with Transformer-based architectures, even a naive implementation on Mamba- and Convolution-based architectures shows to be an efficient way of fine-tuning, making it thereby easily applicable for a variety of tasks. Whilst LoRA outperforming FFT is generally not a novel observation on language modeling \cite{liu2024dora}, the transfer to image classification offers great potential for downstream applicability to further domains.

\subsection{Regularization of VP}\label{sec:regularization}
Applying L1 regularization to either the prompt generation and the Label Mapping leads to almost similar enhancements. Conversely, for L2 regularization, regularizing the Label Mapping is substantially more effective than regularizing the prompt.
The augmentation schemes from Cubuk et al. \yrcite{cubuk2019autoaugment}, which are derived from ImageNet, SVHN, and CIFAR10, increased performance, however to the smallest degree. VMamba also benefits from the utilization of SAM as well as from applying Dropout to the Label Mapping.
\vspace{-2mm}
\begin{table}[h]
\centering
     \begin{sc}
    \begin{tabular}{|l|c|}
        \hline
        Reference   & 67.87 \\
        Dropout     & 68.73 \\
        L1 Prompt   & 69.42 \\
        L1 Mapping  & 69.37 \\
        L2 Prompt   & 68.84 \\
        L2 Mapping  & \textbf{70.31} \\
        Autoaugment & 68.65 \\
        SAM         & 69.31 \\
        \hline
    \end{tabular}
    \end{sc}
    \captionof{table}{\textit{AutoVP performances of VMamba-B using different regularization mechanics. First row reports the unregularized VP performance.}}
    \label{tab:reg}
    \vspace{-6mm}
\end{table}

For VMamba-B on DTD, the tested regularizers improve VP accuracy but leave a gap
to LoRA. Also, no single regularization method can completely close the performance gap between VMamba and ConvNext.
\vspace{-2mm}
\section*{Conclusion}
The VMamba-T ablation on Flowers102 shows that the choice of label mapping
can strongly affect visual prompting performance, highlighting the importance of the adaptation
procedure when assessing a pretrained model. When employing VP, our results show that the selection should consider both its pretraining accuracy and the model architecture, with ConvNext being the best option in our setup. In particular on data scarce tasks, LoRA seems to be the method of choice, outperforming FFT while being lightweight.

\section*{Acknowledgments}
The authors acknowledge the financial support of the Hector Foundation.
\vspace{10mm}
\bibliography{ressources}
\bibliographystyle{icml2025}

\newpage
\appendix
\onecolumn
\section{Benchmarking Visual Prompting}\label{App: AutoVP}

The Experiments from \ref{sec:VP} followed the AutoVP recipe \cite{tsao2023autovp} with:
\begin{itemize}
    \item Learning rates of [0.0001, 0.0003, 0.001, 0.003, 0.01, 0.03, 0.1, 0.3]
    \item 2 different seeds and no regularization applied.
    \item Initial image-scale of 1.0 with prompt scaling enabled
    \item 50 epochs
\end{itemize}
The reported accuracy is the maximum test accuracy over the run.
Runs comparing the different Mapping approaches were conducted with learning rates [0.0001, 0.0003, 0.001, 0.003, 0.01, 0.03, 0.1, 0.3] with one seed.

\begin{table*}[h]
    \centering
    \begin{sc}
    \begin{tabularx}{\columnwidth}{|X|c c c c c c c c| c c |}
    \hline
    \centering{} & \rotatebox{90}{CIFAR10} & \rotatebox{90}{CIFAR100} & \rotatebox{90}{Flowers102} & \rotatebox{90}{GTSRB} & \rotatebox{90}{DTD} & \rotatebox{90}{EuroSAT} & \rotatebox{90}{Food101} & \rotatebox{90}{OxfordPets} & \rotatebox{90}{$\mu$} & \rotatebox{90}{$\sigma$} \\
    \hline
    VMamba-T                & 0.990  & 0.999  & 0.996  & 1.047  & 0.999  & 1.083  & 1.000  & 0.999  & 1.014  & 0.031  \\ 
    MambaOut-T              & 1.000  & 0.999  & 0.999  & 3.674  & 0.997  & 0.691  & 1.254  & 2.275  & 1.486  & 0.938  \\ 
    MambaVision-T           & 0.995  & 0.998  & 1.401  & 3.111  & 2.195  & 0.927  & 1.182  & 2.342  & 1.644  & 0.756  \\ 
    SwinT-T                 & 0.986  & 1.001  & 0.999  & 0.995  & 0.999  & 0.950  & 1.000  & 1.196  & 1.016  & 0.070  \\
    MambaVision-S           & 0.927  & 1.006  & 2.930  & 1.105  & 0.997  & 1.006  & 1.238  & 1.942  & 1.394  & 0.656  \\ 
    \hline
    MambaVision-B           & 1.003  & 1.006  & 1.002  & 3.912  & 2.362  & 0.994  & 1.386  & 1.574  & 1.655  & 0.960  \\ 
    ConvNext-B              & 0.994  & 1.001  & 0.999  & 0.991  & 0.998  & 0.985  & 1.000  & 1.001  & 0.996  & 0.005  \\ 
    VMamba-B                & 1.004  & 1.021  & 0.997  & 0.863  & 0.999  & 0.927  & 1.000  & 0.998  & 0.976  & 0.050  \\
    DeiT-B	                & 1.011	 & 1.000  &	0.998  & 0.792	& 1.000	 & 0.569  &	1.000  & 1.562  & 0.991	 & 0.261  \\
    ViT-B                   & 1.004  & 0.856  & 1.090  & 0.998  & 1.349  & 1.000  & 2.332  & 2.190  & 1.352  & 0.542  \\
    Swin-B                  & 1.003  & 0.997  & 1.001  & 4.798	& 0.999	 & 1.003  & 1.214  & 1.542	& 1.569  & 1.233  \\ \hline
    \end{tabularx}
    \end{sc}
    \caption{\textit{Prompt scaling of the best run for the respective model-task combination}}
    \label{tab:VP Prompt size} 
\end{table*}
Tab. \ref{tab:VP Prompt size} provides detailed results regarding the Prompt resizing for different Model-task combinations. The reported number is the prompt size after the last training epoch w.r.t. the initial instantiation.\\
An interesting observation is that while some tasks generally lead to a stronger resizing (e.g. GTSRB results in an average prompt size of 2.026 times the initial calibration), 5 out of the 11 models observed also reduce the prompt size.

\section{Comparing Knowledge Transfer Methods}\label{App: Reprogramming}
LP and FFT were implemented by adapting \href{https://github.com/kuangliu/pytorch-cifar}{https://github.com/kuangliu/pytorch-cifar}.  The results for VP were obtained using the setup described in \ref{App: AutoVP}.

The models were all evaluated on FFT and LP using the combinations of the learning rates [0.0001, 0.0003, 0.001, 0.003, 0.01, 0.03, 0.1, 0.3] and weight decays [0.0001, 0.0003, 0.001, 0.003, 0.01, 0.03, 0.1, 0.3]. All models were trained for 200 epochs on one seed. 

LoRA was implemented using the \href{https://github.com/huggingface/peft}{\texttt{peft}} \cite{peft} library. 

\texttt{RandomRotation((10))}, \texttt{Colorjitter(brightness=0.05,contrast=0.05,saturation=0.05)} and \texttt{RandomAffine(degrees=5,translate=(0.1,0.1),scale=(0.9,1.1))} were used as input perturbations. 

For LP and FFT, the linear layer of the classification head was replaced to account for the correct number of output classes. For LP, the rest of the model was frozen and only the new classifier was trained.

\section{Regularizing Visual Prompting}\label{App: reg}
The Experiments from \ref{sec:regularization} used the best performing learning rate of 0.001.
\begin{itemize}
    \item Dropout was applied with p-values of [0.05, 0.1, 0.2, 0.4] on the Label Mapping.
    \item L1 and L2 regularization were both run with $\lambda$-values of [0.0001, 0.0003, 0.001, 0.003, 0.01, 0.03, 0.1, 0.3] on either the prompt or the Label Mapping. 
    \item Autoaugment was used by applying the precomputed schemes for ImageNet, SVHN and CIFAR10.
    \item For SAM, $\rho$ of [0.001, 0.003, 0.005, 0.008 and 0.01] were used.
\end{itemize}


\end{document}